\documentclass[10pt]{llncs}

\usepackage{mathtools}
\usepackage{longtable}
\usepackage{pdflscape}
\usepackage{color}
\usepackage{graphicx}
\usepackage[noend]{algorithmic}
\usepackage{algorithm}
\usepackage{array}
\usepackage{amsmath}
\usepackage{amssymb}
\usepackage{microtype}
\usepackage{hyperref}
\usepackage{amsfonts}
\usepackage{cleveref}
\usepackage{subfig}

\title{\LARGE \bf
A Shapelet Transform for Multivariate Time Series Classification
}

\author{Aaron Bostrom and Anthony Bagnall}
\institute{University of East Anglia, Norwich, UK, NR47TJ.}

\begin{document}

\maketitle

\begin{abstract}
Shapelets are phase independent subsequences designed for time series classification. We propose three adaptations to the Shapelet Transform (ST) to capture multivariate features in multivariate time series classification. We create a unified set of data to benchmark our work on, and compare with three other algorithms. We demonstrate that multivariate shapelets are not significantly worse than other state-of-the-art algorithms.
\end{abstract}

\section{INTRODUCTION}
Multivariate time series classification (MTSC) has gained traction in recent years although the majority of work in time series classification (TSC) has focused on the univariate case. Where a single signal for univariate TSC is assigned a class label, multiple signals are recorded for one class with MTSC.

Multivariate time series classification has many practical applications. These can range from medical problems, such as electroencephalogram (EEG), finance, multimedia, human activity recognition (HAR) and gesture recognition. It is commonly claimed that transitioning to multivariate from univariate is trivial (eg~\cite{shokoohi17generalizing}). However, we do not believe this is necessarily true. 

Recently a large experimental analysis of the state of univariate time series classification was conducted \cite{bagnall16bakeoff}. One of the most successful algorithms within that study was the Shapelet Transform (ST) \cite{lines12shapelet,bostrom15binary}. Shapelets are discriminative phase-independent subsequences that have been used in clustering, classification, and early prediction \cite{large2017hesca}.  
The Shapelet Transform was proposed as an improvement to the Shapelet Tree algorithm \cite{ye11shapelets} where the shapelets were used to form the rules within a decision tree. However, it was later shown that using shapelets to perform a data transformation which is then paired with more complex classifiers produces significantly better accuracy \cite{lines12shapelet,hills14shapelet}. More recently, in MTSC, shapelets have been used in forests of randomized shapelet trees \cite{karlsson15forests,karlsson17order}. We wish to investigate whether the simpler transform approach can be adapted to do as well, or better than, tree based approaches. 

We evaluate some of the state-of-the-art multivariate time series classification algorithms, as well as collate datasets from the literature and contribute one of our own. We define three multivariate shapelet algorithms and evaluate a number of classifiers to build a large comparison of algorithms on 22 MTSC datasets. We present a constrained version of ST which can be used on very large problems where enumeration is infeasible. The purpose of this work is to provide a better set of benchmarking to aid improvements in this area of research. We make the data, results and code publicly available~\footnote{http://research.cmp.uea.ac.uk/multivariate\_shapelets/}. These multivariate datasets will form the beginning of the multivariate UEA-UCR dataset archive. 

The first shapelet approach we evaluate is a phase and dimension independent shapelet transform. Secondly, we define an inter-dimensional shapelet method, MST\textsubscript{I} which finds intra-phase dependent subsequences and compare them on there inter-phase independent similarity. Finally we define MST\textsubscript{D} which finds intra-phase dependent subsequences but compares them on their inter-phase dependent similarity. Both of the inter-dimensional shapelet methods are used to form the Multivariate Shapelet Transform. 

The remainder of this paper is organised as follows: in \autoref{sec:background} some of the initial work exploring MTSC is presented.  In \autoref{sec:shapeletbackground} the background and algorithm definitions for the shapelet transform are presented. In \autoref{sec:datasets} the datasets are introduced and these will be used to assess the proposed shapelet algorithms. We describe how the MVMotion datasets were constructed. In \autoref{sec:algorithms} we introduce the three shapelet representations and describe how they fit into the Shapelet Transform framework. In \autoref{sec:evaluation} we conduct a thorough analysis of the shapelet methods against current multivariate time series classification algorithms we have implemented from within the literature. Finally in \autoref{sec:conclusion} the work is concluded and suggest further improvements proposed.

\section{Background}
\label{sec:background}

we define a MTSC dataset as a set of $n$ time series,
\textbf{MT}$= \{MT_1,MT_2,...,MT_n\},$ 
where a single time series $MT_i = \{\{T_{i,1},T_{i,2},...,T_{i,d}\}, c\}$ is a set of $d$ time series with a single shared class label. Each series in a multivariate instance is described as $T_{i,j} = \{t_{i,j,1},t_{i,j,2},...,t_{i,j,m}\}$ where we define the length as $m$ real-ordered values. 

A large volume of research has been conducted on analyzing multivariate time series. The breadth of this work spans from gesture recognition to mining of historical documents and handwriting.
Gesture and human activity recognition is one of the most popular areas of research with in this field \cite{ten07multi,lara13survey,liu09uwave,ermes08detection,kale12impact,ko05online,kela06accelerometer}. Gesture recognition has also been extended to particular activities such as playing  musical instrument \cite{gillian11recognition,tang13extracting}. Multivariate research in the health domain has focused on health records, electroencephalogram (EEG) classification, or balance and mobility sensor data for patients with Parkinsons disease (PD) \cite{mikalsen16learning,gorecki15multivariate,al12using}.
Other domains have considered handwriting classification \cite{bashir08reduced}, similarity between image textures \cite{de08multi} and mining of historical manuscripts \cite{zhu10mother}.

Most of these research domains have focused on using dynamic time warping with a nearest neighbor classifier, mainly because until very recently it was considered the state of the art solution to time series classification \cite{ratanamahatana05three}. 

Various forms of multivariate dynamic time warping have been proposed. The dynamic time warping algorithm is partially modified to consider two difference types of multivariate similarity. These types of features are considered independent and dependent of the dimension, dependent dynamic time warping (DTW\textsubscript{D}) was proposed for use in historical text mining \cite{chen13dtw} and Independent dynamic time warping was later proposed in addition to adaptive dynamic time warping in \cite{shokoohi17generalizing}. 

DTW\textsubscript{D} and DTW\textsubscript{I} are very simple modifications to the DTW algorithm. Given two dimensional multivariate time seires, $Q$ and $C$ which have two dimensions $X$ and $Y$, dDTW\textsubscript{D} finds the shortest path when combining distances inside the warping window (\autoref{eq:DTWD}). DTW\textsubscript{I} calculates individual distances for each series and each dimension, and then combines the distances (\autoref{eq:DTWI}). Adaptive dynamic time warping (DTW\textsubscript{A}) was defined as way to dynamically select which multivariate version of DTW was best suited to the dataset and demonstrated that in the worst-case DTW\textsubscript{A} was no worse than either of its components.

\begin{equation}
\label{eq:DTWD}
DTW_D(Q,C) = DTW({Q_X, Q_Y}, {C_X,C_Y})
\end{equation}

\begin{equation}
\label{eq:DTWI}
DTW_I(Q,C) = DTW(Q_X,C_x) + DTW(Q_Y, C_Y)
\end{equation}

The Shapelet Forest algorithm was presented as one of the first multivariate implementations of shapelets for multivariate time series classification \cite{patri15multivariate,karlsson17order,karlsson15multi}. Grabocka proposed Ultra-fast shapelets, that are learned from the set of all possible shapelets \cite{grabocka16shapelets}.

\section{TSC with Shapelets}
\label{sec:shapeletbackground}
One of shapelets major features on univariate problems is that they can produce easy to understand features, and interpretable results. The aim of this work is to provide the interpretability for multivariate problems whilst maintaining comparable accuracy with other multivariate algorithms, and significantly better accuracy than the univariate algorithms.

\subsection{Shapelet Transform}
The original shapelet transform enumerated all possible shapelets. However, we have found that enumeration is very rarely required and sampling a tiny proportion of the shapelet space does not lead to a significant decrease on accuracy \cite{bostrom16evaluating}.
Algorithm~\ref{alg:binaryShapelet} describes the revised ST algorithm that samples the space of all possible shapelets with the function \texttt{sampleShapelets}. We have explored a range of sampling and heuristic search techniques for finding shapelets. However, none as yet have proved significantly better than simply randomly sampling the shapelet space. Hence, for the datasets that cannot be fully enumerated in a pre specified time, we will randomly sample shapelets from the whole space until the algorithm runs out of time. We define this approach as a contract classifier, and the subsequent definitions will define a contracted Shapelet Transform. We calculate the total number of shapelets in the whole search space, and estimate how long they will take to calculate. Given a fixed run time of one hour, we then calculate the proportion of shapelets it is feasible to evaluate. Each shapelet found is evaluated by sliding the candidate along the other series and calculating the euclidean distance at each position to find the minimum. The euclidean distance is defined in \autoref{eq:euclid} and the sliding window function is defined in \autoref{eq:sDist}. For series $A$ and $B$ of length $m$ is Euclidean distance is defined as
\begin{equation}
\label{eq:euclid}
dist(A,B) = \sqrt{\sum\limits_{i=1}^{m}(a_i - b_i)^2}.
\end{equation}
We define the sliding window function for calculating the distance between a shapelet and a series in \autoref{eq:sDist}. Where $W$ is the set of all subsequences which are the same length as $S$ in $T$:
\begin{equation}
\label{eq:sDist}
sDist(S,T) = min_{w \in W}(dist(S,w)).
\end{equation}

The information gain of a shapelet is calculated by using the set of distances of a shapelet to each series then calculating the best split of the data. This quality value is used to sort the best shapelets in the dataset. In earlier work we changed the shapelet transform to use a binary encoding for class values which was shown to be significantly better on multi-class problems \cite{bostrom15binary}.

The $k$ best shapelets are used to create a data transformation. By using the distance between the shapelet and each series, we form a $k$ by $n$ feature matrix. One of the main advantages of using transformed data is that we can use any classifier, as opposed to using only a decision tree as per the original definition of the algorithm. The classifier typically used in conjunction with ST is the heterogeneous ensemble of simple classification algorithms (HESCA) \cite{lines17hive,large2017hesca}. 

The algorithm is simplified in comparison to the implementation for ease of explanation. This description omits the class balancing, binary shapelet methods, and the current state of the art calculation heuristics.

%TODO: write this again. 
\begin{algorithm}
\caption{ShapeletTransform(\textbf{T}, $min$, $max$, $k$, $totalShapelets$)}
\label{alg:binaryShapelet}
\begin{algorithmic}[1]
\REQUIRE A list of time series \textbf{T}, $min$ and $max$ length shapelet to search for and $k$, the maximum number of shapelets to find
\ENSURE A list of $k$ Shapelets
\STATE $shapeletCount = 0$
	\WHILE{$shapeletCount \le totalShapelets$}
      \STATE $S = extractRandomShapelet($\textbf{T},$min, max)$
      \STATE $shapeletCount = shapeletCount + 1$
      \STATE $D_S \leftarrow findDistances(S, $\textbf{T})
      \STATE $quality \leftarrow assessCandidate(S, D_S)$
      \STATE $shapelets.add(S, quality)$
  	\ENDWHILE
\STATE $kShapelets = extractKBest(shapelets)$
\RETURN $kShapelets$
\end{algorithmic}
\end{algorithm}

\section{Datasets}
\label{sec:datasets}
In \autoref{fig:datasets} we present the list and the properties of the multivariate datasets we have collated from the literature. The datasets have a range of different sizes, number of instances, length of the series, the number of series and finally number of classes. To simplify and reduce the need for extensive dataset knowledge we have reduced some problems into sub problems. This is most notable with the AALTD problems. These were originally from a challenge dataset produced for the ECML/PKDD Workshop on Advanced Analytics and Learning on Temporal Data (AALTD).
The aim being to classify six different gestures using eight spatial sensors placed on a person, resulting in 3 dimensional movement information for each sensor. We split the dataset into a separate classification problem for each sensor. However, this will have the effect of artificially making the problems more difficult as class discriminating information may not be contained in a particular sensory dataset, or could even be across multiple sensors.

Most of the datasets have been used in the literature, our aim is to unify these under a common framework using the ARFF format in Weka \cite{Weka}. The following datasets were extracted and converted:
AALTD; ArabicDigit \cite{hammami09tree}; Japanese vowels \cite{kudo99multidimensional}; Cricket, Handwriting, ArticularyWord \cite{shokoohi17generalizing}; PEMS \cite{cuturi11fast}; PenDigits \cite{alimouglu01combining}; UWaveGesture \cite{liu09uwave}; Epilepsy \cite{villar13human}.

\begin{figure}
\caption{An example of the four classes for both Accelerometer data from the MVMotion dataset.}
\begin{center}
\begin{tabular}{cc}
\subfloat[Accelerometer MVMotion datasets]{
	\includegraphics[width=\linewidth/2]{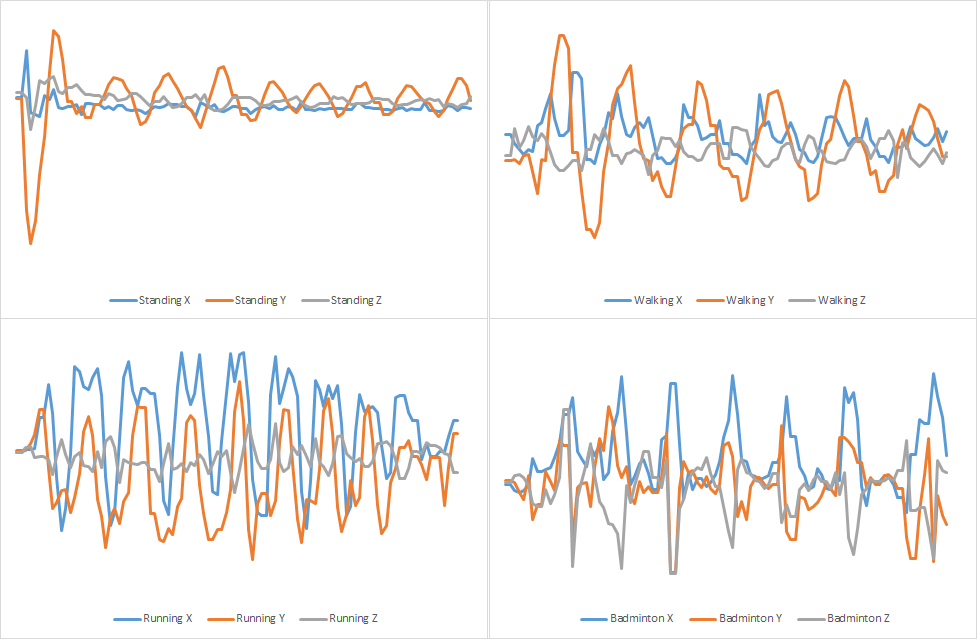}
    \label{fig:MVMotionA}
}
\subfloat[Gyroscope MVMotion dataset]{
	\includegraphics[width=\linewidth/2]{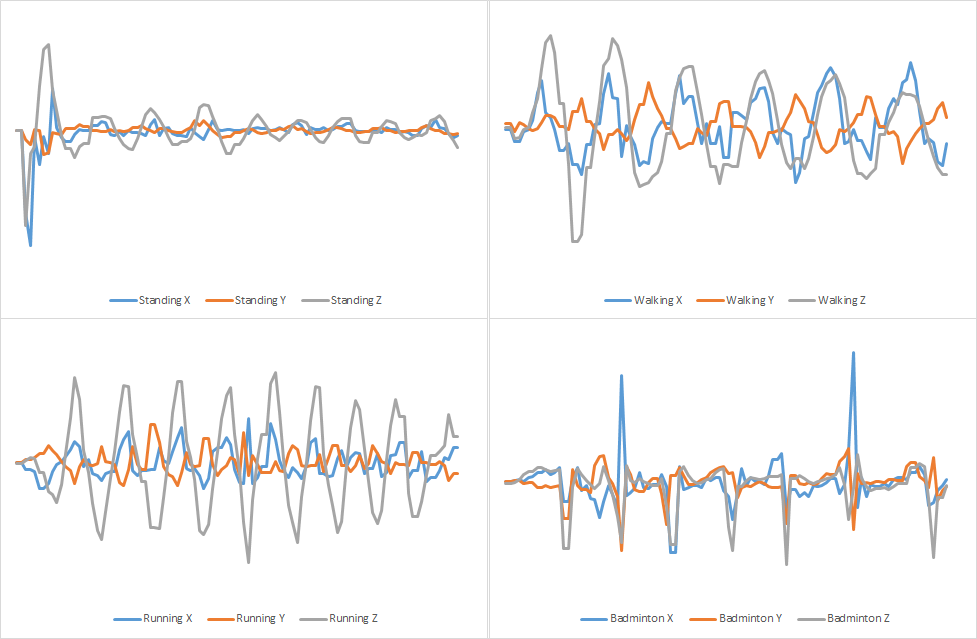}
    \label{fig:MVMotionG}
}
\end{tabular}
\end{center}
\end{figure}

The final dataset is MVMotion. There are three variants: MVMotionA; MVMotionG; and MVMotionAG. The MVMotion datasets are collected from a 3D accelerometer and a 3D gyroscope on a mobile device during a particular set of activities. The general type of problem is Human Activity Recognition (HAR) and is similar in concept to the Epilepsy dataset. All MVMotion datasets consist of four classes, which are walking, resting, running and badminton. Participants were required to record motion a total of five times, and the data is sampled once every tenth of a second, for a ten second period. We demonstrate an example of each of the classes, for accelerometer data in Figure \autoref{fig:MVMotionA} and for the gyroscope data in Figure \autoref{fig:MVMotionG}. The three datasets are then constructed, MVMotionA is X,Y,Z accelerometer data, MVMotionG is X,Y,Z gyroscope data, and MVMotionAG is both sets combined, to form a six dimensional problem. 

\begin{figure}
\begin{center}
\tiny
\caption{A list of the datasets in the multivariate time series archive. Number of instances is denoted by n, number of dimensions is denoted by d, length of series is denoted by m, and number of classes is denoted by c}.
\label{fig:datasets}
\begin{tabular}{cc}
\begin{tabular}{|c|c|c|c|c|}
\hline
datasets & n & d & m & c\\
\hline
AALTD\_0 & 90 & 3 & 52 & 6\\
AALTD\_1 & 90 & 3 & 52 & 6\\
AALTD\_2 & 90 & 3 & 52 & 6\\
AALTD\_3 & 90 & 3 & 52 & 6\\
AALTD\_4 & 90 & 3 & 52 & 6\\
AALTD\_5 & 90 & 3 & 52 & 6\\
AALTD\_6 & 90 & 3 & 52 & 6\\
AALTD\_7 & 90 & 3 & 52 & 6\\
ArabicDigit & 6599 & 13 & 94 & 10\\
AWordLL & 275 & 3 & 145 & 25\\
AWordT1 & 275 & 3 & 145 & 25\\
AWordUL & 275 & 3 & 145 & 25\\
\hline
\end{tabular}

\begin{tabular}{|c|c|c|c|c|}
\hline
datasets & n & d & m & c\\
\hline
CricketLeft & 84 & 3 & 1198 & 12\\
CricketRight & 84 & 3 & 1198 & 12\\
HandwritingA & 150 & 3 & 153 & 26\\
HandwritingG & 500 & 3 & 153 & 26\\
JapaneseVowels & 270 & 12 & 30 & 9\\
MVMotionA & 40 & 3 & 101 & 4\\
MVMotionAG & 40 & 6 & 101 & 4\\
MVMotionG & 40 & 3 & 101 & 4\\
PEMS & 267 & 144 & 964 & 7\\
PenDigits & 7494 & 2 & 9 & 10\\
UWaveGesture & 120 & 3 & 316 & 8\\
VillarData & 137 & 3 & 207 & 4\\
\hline
\end{tabular}
\end{tabular}
\end{center}
\end{figure}

\section{Multivariate Shapelet Transform} 
\label{sec:algorithms}
\begin{figure}
\begin{center}
\caption{An example of extracting a single shapelet from a many dimensional series, and comparing it to a different series of the same dimension}.
\label{fig:indShapelets}
\begin{tabular}{c}
\subfloat[Shapelet extraction]{
       \includegraphics[width=\linewidth/3]{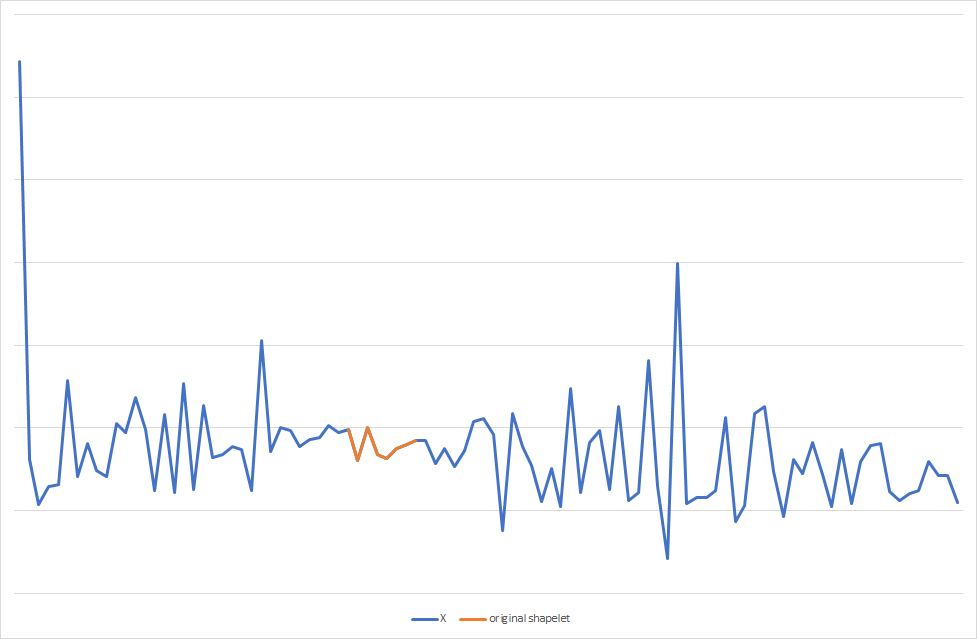}
}
\subfloat[Shapelet matching]{
       \includegraphics[width=\linewidth/3]{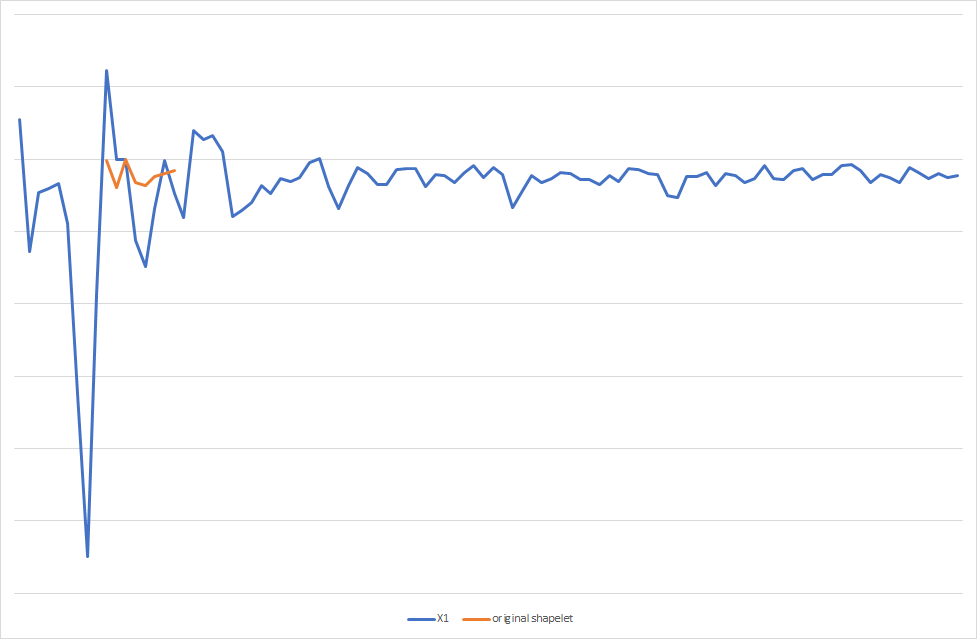}
}
\subfloat[Normalised distance]{
       \includegraphics[width=\linewidth/3]{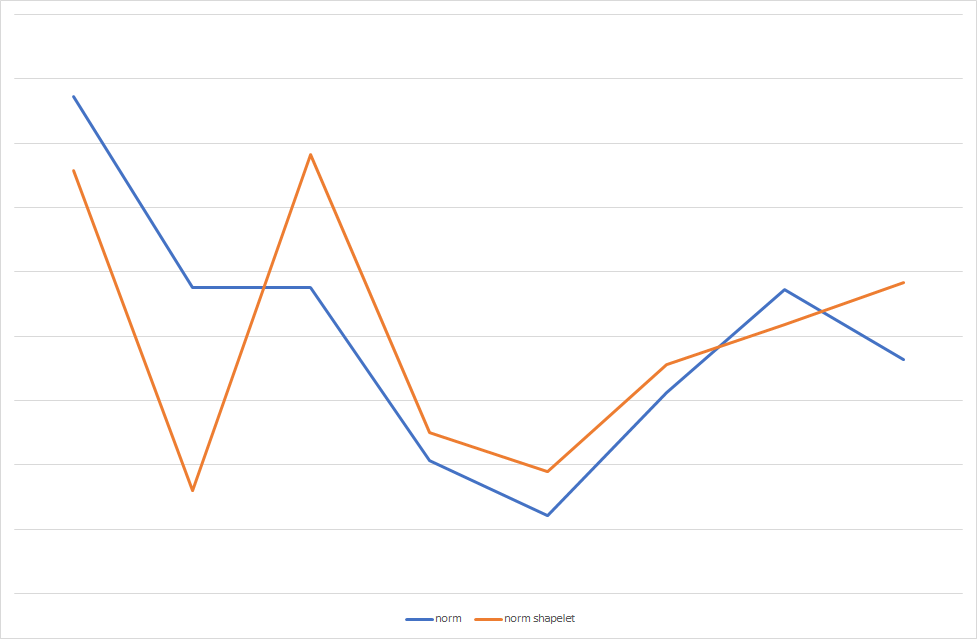}
}
\end{tabular}
\end{center}
\end{figure}

\begin{figure}
\begin{center}
\caption{An example of extracting a Shapelet\textsubscript{D} from a many dimensional series, and comparing it to a different series}.
\label{fig:ShapeletsD}
\begin{tabular}{c}     
\subfloat[Shapelet extraction]{
       \includegraphics[width=\linewidth/3]{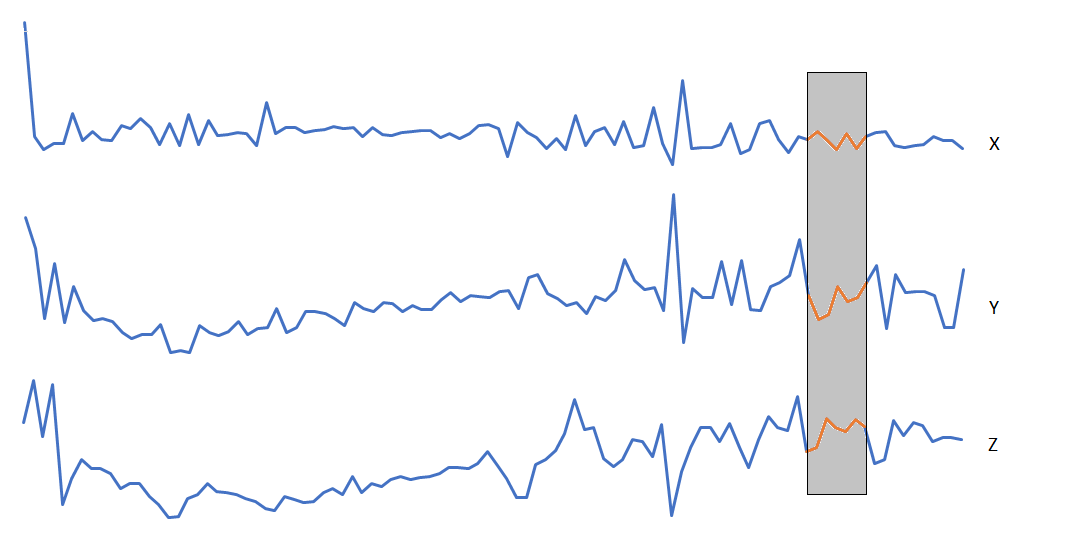}
}
\subfloat[Shapelet matching]{
       \includegraphics[width=\linewidth/3]{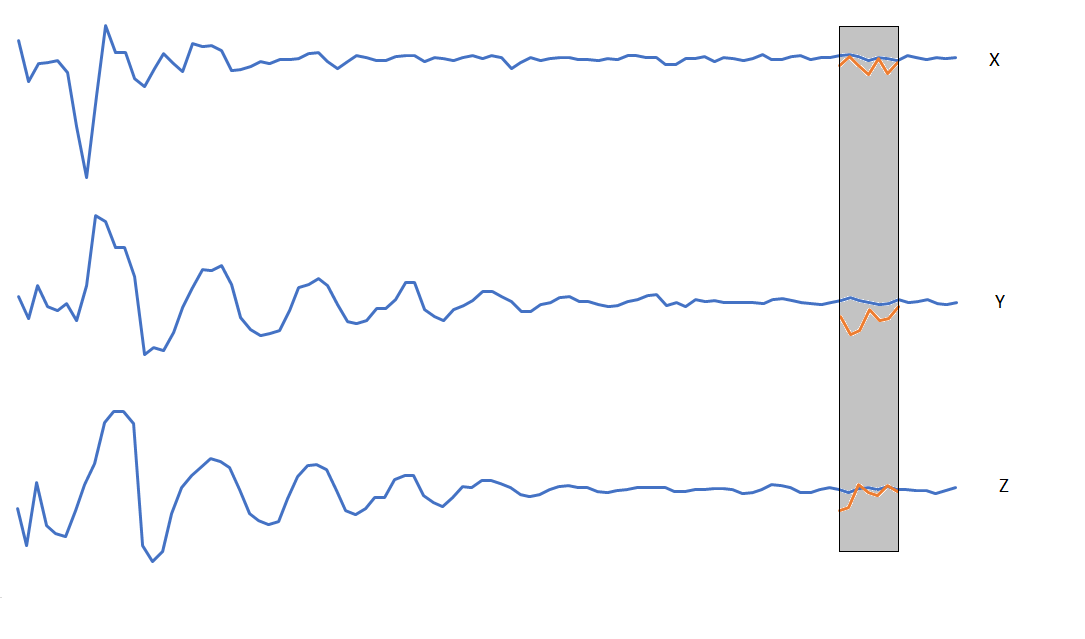}
}
\subfloat[Normalised distance]{
       \includegraphics[width=\linewidth/3]{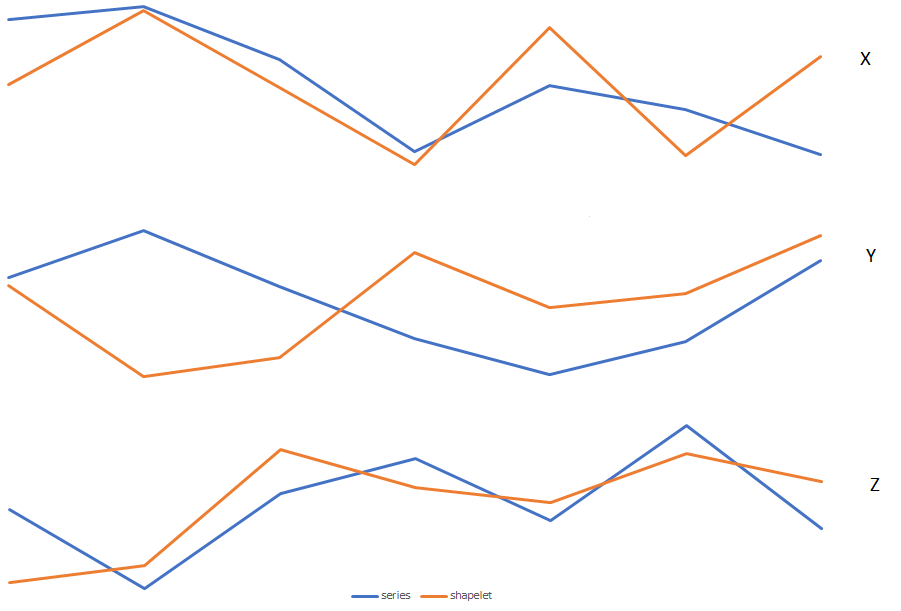}
}
\end{tabular}
\end{center}
\end{figure}

\begin{figure}
\begin{center}
\caption{We present an illustrative example of extracting a Shapelet\textsubscript{I} from a many dimensional series, and comparing it to a different series}.
\label{fig:ShapeletsI}
\begin{tabular}{c}
\subfloat[Shapelet extraction]{
       \includegraphics[width=\linewidth/3]{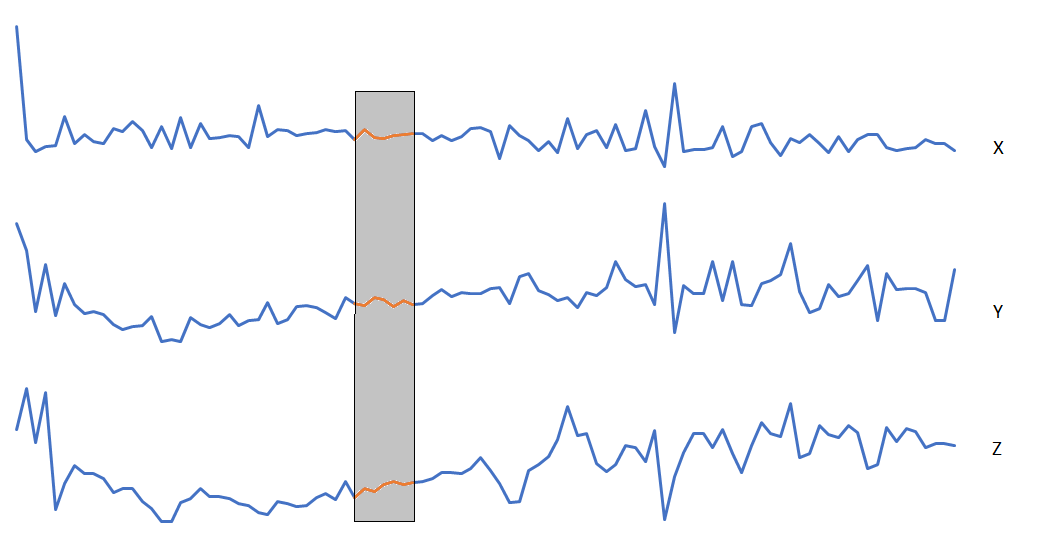}
}
\subfloat[Shapelet matching]{
       \includegraphics[width=\linewidth/3]{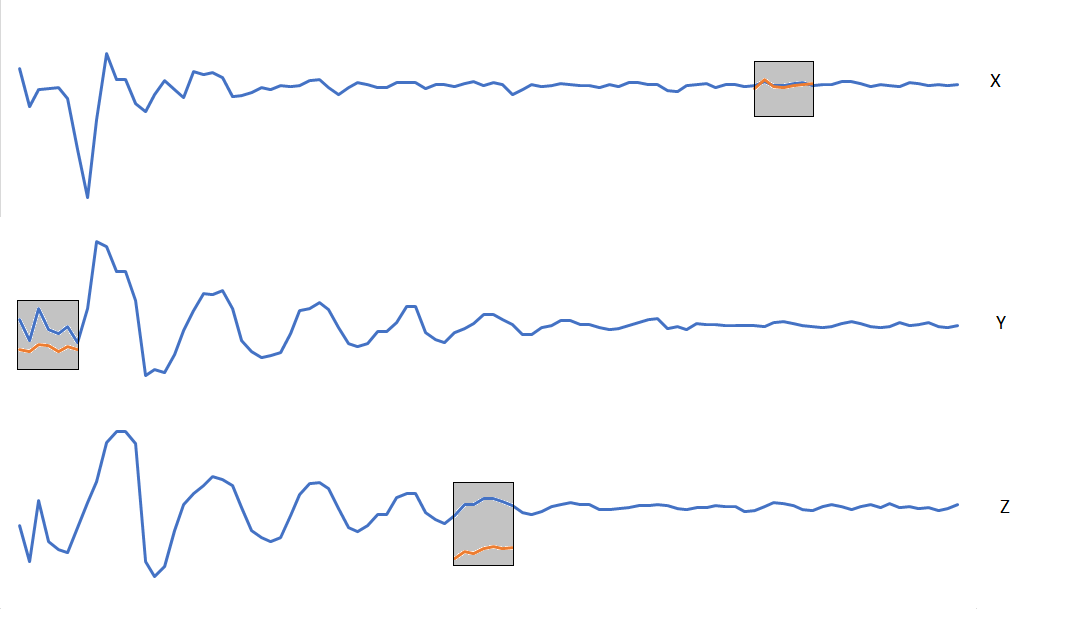}
}
\subfloat[Normalised distance]{
       \includegraphics[width=\linewidth/3]{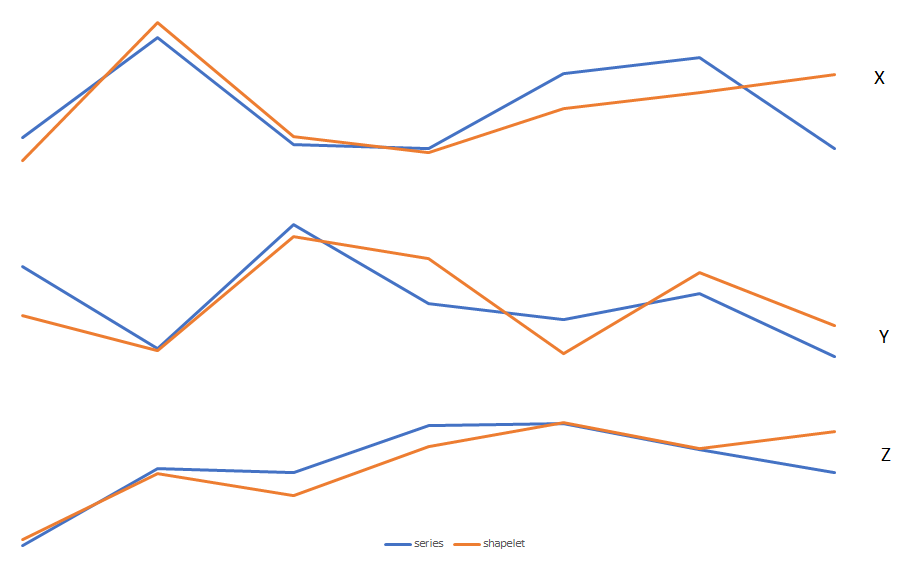}
}
\end{tabular}
\end{center}
\end{figure}

In this section we describe the three shapelet methods developed for the Multivariate Shapelet Transform. 
To illustrate the different methods, a series of figures have been created from a common dataset, demonstrating how all three methods find and extract different features, and how they calculate shapelet to series distance. The MVMotionA dataset is used in Figures~\ref{fig:indShapelets}, \ref{fig:ShapeletsD}, and \ref{fig:ShapeletsI}. In all three figures the left panel is the multivariate series shapelet extraction. In the middle left panel we demonstrate finding the minimum distance point whilst using the sliding window function. In \autoref{eq:sDist} we refer to this sliding window function as $sDist$. In the right panel, we show the exact comparison of the shapelet to the region in the multivariate series where both have been z-normalised. 

% \begin{figure}[!ht]
% \begin{center}
% \caption{A single XYZ series from the MVMotion dataset, class 0}.
% \label{fig:MVMotion}
% \begin{tabular}{c}
%        \includegraphics[width=\linewidth/2]{diagrams/MVMotion_A_series0_XYZ.png} 
% \end{tabular}
% \end{center}
% \end{figure}

\subsection{Independent Shapelets}
\label{ssec:indShapelets}
The first multivariate shapelet method is called Independent Shapelets. This algorithm finds single channel shapelets from any dimension. It then assesses the shapelets quality against the other series via sliding the shapelet along the same channel in the multivariate series. Once the $k$ best shapelets have been found, they are used to transform the original dataset. Using the same distance method, we can transform the multivariate dataset in a $k$ by $n$ matrix, where find the respective distance of the shapelets to each series.  

The motivation for this method is that in some multivariate datasets the class defining feature may occur in only one channel, and it could even be independent of channel. The shape of the feature is the class identifier not its position or channel. We believe this method is most suited if you have multiple channels from different types of data recording where the dimensions are unrelated. One of the datasets we present MVMotionAG contains three channels of accelerometer data, and three channels of rotational(Gimbal) data. In some of the activity recognition the rotational data is completely independent of the movement information. 

\subsection{Multidimensional Dependent Shapelets}
The second multivariate shapelet method is called Shapelet\textsubscript{D}. This method extracts multi-dimensional shapelets, that are then compared to the other multivariate series, maintaining the phase across channels. 

Once we have extracted a multi-dimension shapelet we slide each channel, simultaneously along each channel of all the other series, finding the minimum distance for each multivariate series with respect to the shapelet. Calculating a single distance for the multi-dimensional shapelet against all the other series means the algorithm behaves identically to the univariate Shapelet Transform,   including calculating Information Gain, and transforming the data.

The main aspect of the Shapelet\textsubscript{D} algorithm is that the minimum distance for a multivariate series and a multi-dimensional shapelet is that the position of best match is maintained across the channels. The motivation for this method is that for gesture recognition where a particular gesture is performed, all the channels (X,Y and Z) should have information about this event at the same point, but that the phase independence of shapelets means the information can be captured even though it can occur at any time interval. 
\label{ssec:ShapeletsD}

\subsection{Multidimensional Independent Shapelets}
\label{ssec:ShapeletsI}
The third multivariate shapelet method is called Shapelet\textsubscript{I}. This method is similar to Shapelet\textsubscript{D} which extracts multi-dimensional shapelets. These shapelets are then compared the other multivariate series. Each shapelet within the multi-dimensional shapelet finds the minimum distance to its respective channel independent of the other calculations. Shapelet\textsubscript{D} could be considered a special case of the Shapelet\textsubscript{I} algorithm where the best independent phase matches coincide to be the same. 

The motivation for this method is that we believe whilst shapelet extracted is dependent on the features being in phase, the places where they occur in other series could be independent of one another.

\section{Evaluation}
\label{sec:evaluation}
The experimental setup follows the same approach outlined in \cite{bagnall16bakeoff}. We perform 100 fold resampling on the data. For each algorithm presented we have performed 2,400 experiments. The data presented in the tables are the mean average accuracy across the 100 folds. The critical difference diagrams are calculated from these mean averages. Resampling the datasets aids in identifying overall problem performance. When creating and processing these datasets from the literature and from our own data collection, the train and test splits are originally assigned randomly, thus on a single split the problem could be made arbitrarily hard. Some of the classifiers we use contain a random element, and so our experiments are all seeded to ensure reproducibility. 

\begin{figure}
\begin{center}
\caption{A critical difference diagram demonstrating the the effectiveness of the ensembling and concatenating approaches.}.
\label{fig:cd_simple}
\begin{tabular}{c}
       \includegraphics[width=\linewidth/2]{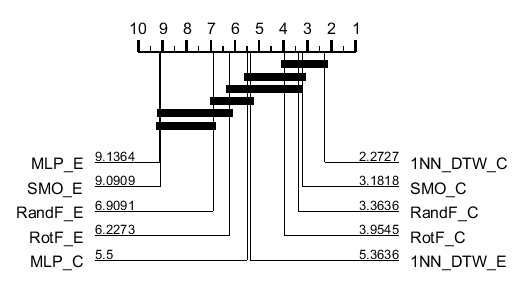}
\end{tabular}
\end{center}
\end{figure}

We initially present the results for the univariate methods for both, concatenation and dimensional ensembling. These algorithms are RotationForest(RotF), RandomForest (RandF), Support Vector Machine, Multi-Layer Perceptron and 1nearest neighbour with dynamic time warping. We use the notation \_C to denote concatenation, and \_E to denote ensembled across dimensions. These results show how concatenating dimensions into a single univariate series is superior to dimensional ensembling. However there are a few cases where the ensemble approach for DTW out performs concatenation. We believe given a cross-validated training approach to weight the respective ensemble dimensions, the ensembled method would be more robust. However the scale, and time requirements to cross-validate on these datasets is out of the scope of this work. The critical difference diagram \autoref{fig:cd_simple} shows there is no significant difference between DTW, the SVM, or random forest and rotation forest.

\subsection{Shapelets}
We present the results for our three multivariate shapelet methods on 17 datasets. Only 16 of the 22 datasets were able to complete full enumerations within a one week runtime limit on our HPC and the results from these are presented in \autoref{fig:initResults} in the second portion of the table we have also included a fixed one hour run time evaluation of the shapelet space for five additional datasets. 

\begin{figure}
\begin{center}
\caption{Two critical difference diagrams demonstrating the effectiveness of the three shapelet transforms and the three multivariate DTW algorithms with full enumeration and one hour random shapelet selection}.
\begin{tabular}{cc}
\subfloat[Full enumeration with 17 datasets]{
       \includegraphics[width=\linewidth/2]{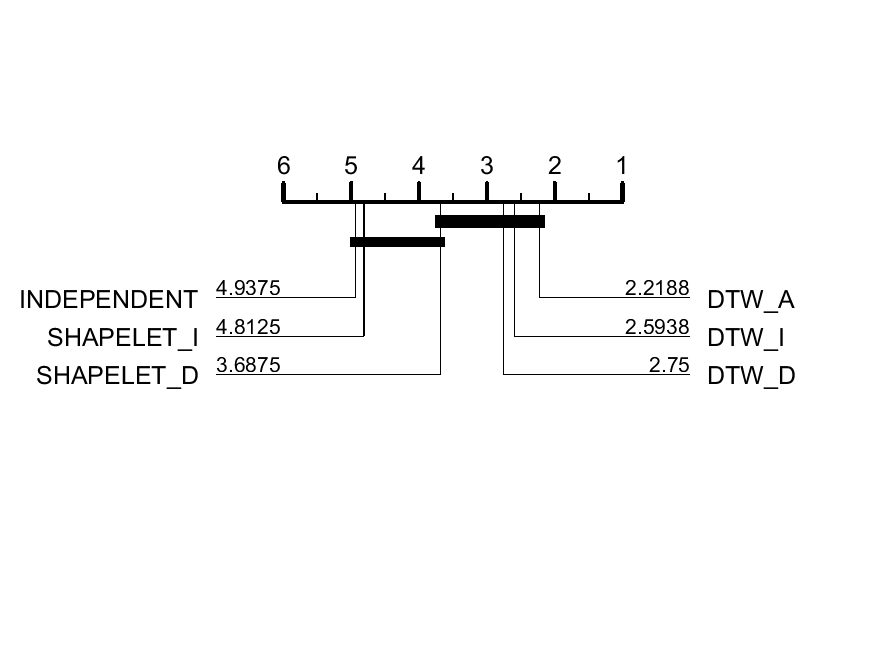}
       \label{fig:cd_shapelet}
       }
\subfloat[One hour random shapelets with 21 datasets]{
       \includegraphics[width=\linewidth/2]{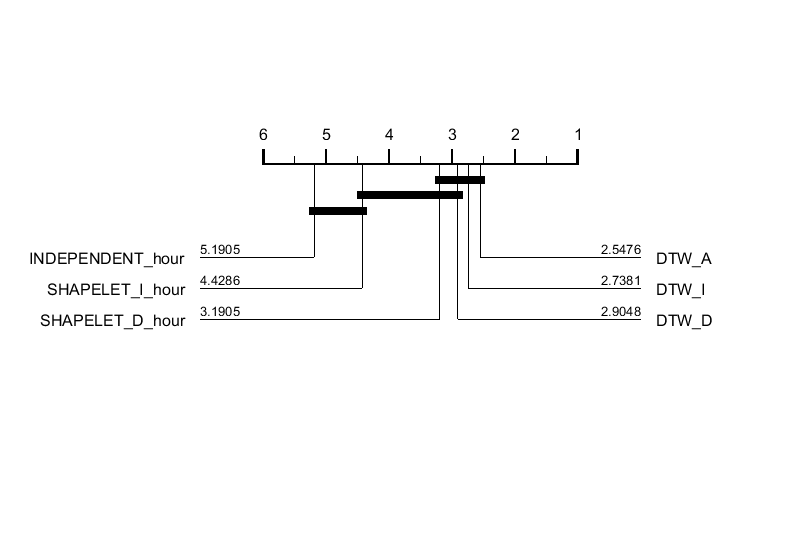}
       \label{fig:cd_shapelet_hour}
       }
\end{tabular}
\end{center}
\end{figure}

Initially we compare the shapelet methods across the 16 datasets over 100 fold resampling and show that Shapelet\textsubscript{D} is not significantly worse than any of the multivariate DTW approaches, illustrated in \autoref{fig:cd_shapelet}. We did multiple pairwise tests on each classifier, and found that Shapelet\textsubscript{D} on a Wilcoxon signed rank test, and the student t-test is not significantly worse than any of the DTW approaches. 

In previous work we have looked at heuristics for constraining the run time of the shapelet transform. A large portion of the univariate problems in the UEA-TSC archive are intractable to fully enumerate, by constraining the number of shapelets that are considered we can fix the run time of the algorithm. We constrained all the datasets to a one hour run time, where necessary. This meant we were able to evaluate an additional five datasets. From these results we find that Shapelet\textsubscript{D} is still not significantly worse than any of the multivariate DTW approaches on 21 datasets all with 100 fold resampling, this is illustrated in \autoref{fig:cd_shapelet_hour}. We performed multiple pairwise tests and show that Shapelet\textsubscript{D} with a constrained runtime of one hour is not significantly worse than any of the three DTW multivariate methods and is not significantly worse than a full enumeration. 

The Shapelet\textsubscript{I} method was significantly worse than the DTW\textsubscript{A} and DTW\textsubscript{I} but was not significantly worse than DTW\textsubscript{D} on a Wilcoxon sign ranked test. Finally we found that the independent shapelet method was significantly worse than all multivariate DTW methods as well as Shapelet\textsubscript{D} and was not significantly worse than Shapelet\textsubscript{I}. Based on these findings Shapelet\textsubscript{D} is an alternative approach to multivariate time series classification.

\begin{figure}
\begin{center}
\caption{}.
\tiny
\label{fig:initResults}
\begin{tabular}{|c|c|c|c|c|c|c|}
\hline
datasets & SHAPELET\_D & INDEPENDENT & SHAPELET\_I & DTW\_A & DTW\_D & DTW\_I\\
\hline
AALTD\_0 & 0.646 (0.04) & 0.583 (0.06) & 0.569 (0.05) & 0.664 (0.03) & \textbf{0.681} (0.03) & 0.649 (0.03)\\
AALTD\_1 & 0.792 (0.03) & 0.725 (0.04) & 0.744 (0.04) & 0.805 (0.03) & 0.804 (0.03) & \textbf{0.809} (0.03)\\
AALTD\_2 & 0.608 (0.04) & 0.529 (0.05) & 0.557 (0.04) & 0.667 (0.04) & \textbf{0.675 (0.03)} & 0.671 (0.04)\\
AALTD\_3 & 0.661 (0.05) & 0.61 (0.05) & 0.656 (0.05) & \textbf{0.684} (0.04) & 0.68 (0.04) & 0.683 (0.04)\\
AALTD\_4 & 0.624 (0.04) & 0.576 (0.04) & 0.619 (0.04) & 0.657 (0.04) & \textbf{0.667} (0.04) & \textbf{0.667} (0.04)\\
AALTD\_5 & 0.767 (0.04) & 0.735 (0.04) & 0.735 (0.04) & 0.789 (0.04) & \textbf{0.797} (0.04) & 0.777 (0.04)\\
AALTD\_6 & 0.617 (0.04) & 0.542 (0.05) & 0.461 (0.06) & 0.654 (0.04) & \textbf{0.671} (0.03) & 0.639 (0.03)\\
AALTD\_7 & \textbf{0.796} (0.04) & 0.739 (0.04) & 0.746 (0.04) & 0.791 (0.03) & 0.784 (0.03) & 0.791 (0.03)\\
ArticularyWordLL & 0.856 (0.02) & 0.828 (0.02) & \textbf{0.865} (0.02) & 0.84 (0.02) & 0.843 (0.02) & 0.83 (0.02)\\
ArticularyWordT1 & 0.923 (0.02) & 0.901 (0.02) & 0.894 (0.02) & 0.921 (0.01) & \textbf{0.924} (0.01) & 0.908 (0.01)\\
ArticularyWordUL & \textbf{0.811} (0.03) & 0.718 (0.03) & 0.829 (0.03) & 0.741 (0.02) & 0.719 (0.02) & 0.749 (0.02)\\
HandwritingA & 0.481 (0.03) & 0.442 (0.03) & 0.426 (0.03) & 0.601 (0.03) & \textbf{0.609} (0.02) & 0.48 (0.02)\\
JapaneseVowels & 0.887 (0.02) & 0.808 (0.03) & 0.88 (0.02) & 0.957 (0.01) & 0.955 (0.01) & \textbf{0.959} (0.01)\\
MVMotionA & \textbf{0.979} (0.02) & 0.956 (0.02) & 0.963 (0.03) & 0.912 (0.05) & 0.77 (0.04) & 0.912 (0.05)\\
MVMotionAG & 0.984 (0.02) & 0.953 (0.03) & 0.961 (0.03) & \textbf{0.999} (0)    & 0.951 (0.04) & \textbf{0.999} (0)   \\
MVMotionG & 0.936 (0.04) & 0.939 (0.03) & 0.933 (0.04) & \textbf{0.996} (0.01) & 0.917 (0.04) & \textbf{0.996} (0.01)\\
\hline
\hline
CricketLeft & 0.92 (0.03) & 0.819 (0.04) & 0.869 (0.03) & \textbf{0.927} (0.02) & 0.933 (0.02) & 0.887 (0.02)\\
CricketRight & 0.935 (0.02) & 0.93 (0.03) & 0.93 (0.03) & 0.939 (0.03) & 0.924 (0.03) & \textbf{0.945} (0.03)\\
HandwritingG & 0.84 (0.01) & 0.711 (0.1) & 0.769 (0.11) & 0.861 (0.05) & \textbf{0.863} (0.05) & 0.785 (0.04)\\
UWaveGesture & 0.898 (0.02) & 0.868 (0.02) & 0.862 (0.02) & 0.919 (0.01) & \textbf{0.925} (0.01) & 0.909 (0.01)\\
VillarData & 0.969 (0.01) & 0.978 (0.02) & \textbf{0.981} (0.01) & 0.965 (0.01) & 0.957 (0.02) & 0.969 (0.01)\\
\hline
\end{tabular}
\end{center}
\end{figure}

\section{Conclusion}
\label{sec:conclusion}
In conclusion we have presented three new shapelet algorithms, one of which is not significantly worse than the current state-of-the-art in multivariate time series classification even when the search space is constrained. We have provided a unified set of datasets, under a common framework and file format. We provided some simple benchmarks of standard classifiers on these datasets to enable further research and comparison alongside the multivariate algorithms. 

In future work we hope to be able to provide significantly better accuracy for shapelets, by assessing the way we measure the quality of multivariate shapelets, or by considering alternative distance measures. 

We would also like to implement more of the multivariate algorithms from the literature under a common Java framework in WEKA to enable further analysis.

\bibliographystyle{ieeetr}
\bibliography{TSCMaster,MultivariateTSC}

\end{document}